# AGENT-BASED SIMULATION OF PEDESTRIANS' EARTHQUAKE EVACUATION; APPLICATION TO BEIRUT, LEBANON

R. Iskandar[(1)], K. Allaw[(2)], J. Dugdale[(3)], E. Beck[(4)], J. Adjizian-Gérard[(5)], C. Cornou[(6)], J. Harb[(7)], P. Lacroix[(8)], N. Badaro-Saliba[(9)], S. Cartier[(10)], R. Zaarour[(11)]

[(1)] *PhD candidate, Univ. Grenoble Alps, ISTerre, Pacte, LIG, rouba.iskandar@univ-grenoble-alpes.fr*
[(2)] *PhD candidate, Saint-Joseph University, allaw.kamel@hotmail.com*
[(3)] *Associate professor, Univ. Grenoble Alps, LIG, julie.dugdale@univ-grenoble-alpes.fr*
[(4)] *Associate professor, Univ. Grenoble Alps, Pacte, elise.beck@univ-grenoble-alpes.fr*
[(5)] *Professor, Saint-Joseph University, jocelyne.gerard@usj.edu.lb*
[(6)] *Researcher, Univ. Grenoble Alps, ISTerre, cecile.cornou@univ-grenoble-alpes.fr*
[(7)] *Professor, Notre Dame University, jharb@ndu.edu.lb*
[(8)] *Researcher, Univ. Grenoble Alps, ISTerre, pascal.lacroix@univ-grenoble-alpes.fr*
[(9)] *Associate professor, Saint-Joseph University, nada.saliba@usj.edu.lb*
[(10)] *Researcher, Univ. Grenoble Alps, Pacte, stephane.cartier@univ-grenoble-alpes.fr*
[(11)] *Professor, Saint-Joseph University, rita.zaarour@usj.edu.lb*

## *Abstract*

Most seismic risk assessment methods focus on estimating the damages to the built environment and the consequent socio-economic losses without fully taking into account the social aspect of risk. Yet, human behaviour is a key element in predicting the human impact of an earthquake, therefore, it is important to include it in quantitative risk assessment studies. In this study, an interdisciplinary approach simulating pedestrians' evacuation during earthquakes at the city scale is developed using an agent-based model. The model integrates the seismic hazard, the physical vulnerability as well as individuals' behaviours and mobility. The simulator is applied to the case of Beirut, Lebanon.

Lebanon is at the heart of the Levant fault system that has generated several Mw>7 earthquakes, the latest being in 1759. It is one of the countries with the highest seismic risk in the Mediterranean region. This is due to the high seismic vulnerability of the buildings due to the absence of mandatory seismic regulation until 2012, the high level of urbanization, and the lack of adequate spatial planning and risk prevention policies. Beirut as the main residential, economic and institutional hub of Lebanon is densely populated. To accommodate the growing need for urban development, constructions have almost taken over all of the green areas of the city; squares and gardens are disappearing to give place to skyscrapers. However, open spaces are safe places to shelter, away from debris, and therefore play an essential role in earthquake evacuation. Despite the massive urbanization, there are a few open spaces but locked gates and other types of anthropogenic barriers often limit their access.

To simulate this complex context, pedestrians' evacuation simulations are run in a highly realistic spatial environment implemented in GAMA [1]. Previous data concerning soil and buildings in Beirut [2, 3] are complemented by new geographic data extracted from high-resolution Pleiades satellite images. The seismic loading is defined as a peak ground acceleration of 0.3g, as stated in Lebanese seismic regulations. Building damages are estimated using an artificial neural network trained to predict the mean damage [4] based on the seismic loading as well as the soil and building vibrational properties [5]. Moreover, the quantity and the footprint of the generated debris around each building are also estimated and included in the model. We simulate how topography, buildings, debris, and access to open spaces, affect individuals' mobility. Two city configurations are implemented: 1. Open spaces are accessible without any barriers; 2. Access to some open spaces is blocked. The first simulation results show that while 52% of the population is able to arrive to an open space within 5 minutes after an earthquake, this number is reduced to 39% when one of the open spaces is locked. These results show that the presence of accessible open spaces in a city and their proximity to the residential buildings is a crucial factor for ensuring people's safety when an earthquake occurs.

## 1. Introduction

Earthquakes are sudden onset events that often cause damages to the built environment, social and economic losses, and disruptions to daily life. Urban seismic risk assessment methodologies have taken place worldwide at different geographic levels to assess the potential seismic risk in an area of interest. The main objective of these methods is to inform engineers and decision-makers on the potential threat to their communities, and to help improve their awareness and preparedness to face a future crisis.

Traditional earthquake risk assessment methods in the engineering community often focus on estimating the damages to the built-environment. In these studies, earthquake risk is defined by: the hazard, i.e. the earthquake activity in the region and the geological conditions; the exposure, i.e. the building inventory and its occupancy criteria; and the vulnerability, i.e. the characteristics of the building stock [6]. On the other hand, in the disaster research field, definitions of risk and vulnerability that include underlying social conditions have emerged [7, 8]. This shift in perspective was also reflected in urban seismic risk assessment studies. Interdisciplinary urban seismic risk assessment based on holistic approaches that incorporate societal factors into the assessment have been explored [9, 10].

Human behaviour during crisis is still rarely included in urban seismic risk assessment studies. Yet, the actions taken by individuals during and immediately after an earthquake highly affect their exposure and vulnerability [11]. In an urban area where the landscape is modified by an earthquake, the interactions between the individuals and their surroundings are a key element in determining their safety [12]. Additionally, the availability of open spaces, i.e. areas open to the public that can serve as immediate evacuation destinations, and their ease of access, have a key role in increasing an urban area's ability to adapt and cope with the effects of an earthquake [13, 14]. Urban seismic risk assessments that consider human behaviour and people's interaction with their environment in an earthquake, contribute to the development of improved of mitigation strategies and prevention plans.

Simulating social situations can be achieved though agent-based models [15]. These models can recreate the emergence of complex behaviours through the interaction of agents implemented by simple rules. Moreover, they also allow dynamic simulations in a highly realistic spatial context. Agent-based models have been widely used for recreating natural hazards crisis scenarios, such as floods [16], tsunamis [17] and bushfires [18]. There are also few applications for pedestrian evacuation in earthquake scenarios [19, 20].

This paper proposes an agent-based model of pedestrians' evacuation in an urban environment that has been affected by an earthquake. The particularity of this model is that it includes realistic estimates of damages and debris, in addition to pedestrians' mobility patterns and the constraints that affect them. The objective is to simulate pedestrians' evacuation to open spaces during and a few minutes after an earthquake, and to test the effect of the accessibility of open spaces on the pedestrians' arrival to safe areas. The model is implemented in GAMA and the simulations are applied to the case of Beirut, Lebanon.

The paper is organized as follows: Section 2 describes the conceptualization of the model. Section 3 details the data collection procedure for the application on Beirut. Section 4 describes the implementation of the model and the simulation results.

## 2. Model design

The model described in this section is a model for pedestrians' evacuation in a city during and immediately after an earthquake.

### 2.1 Conceptual model

The evacuation takes place from the individuals' location at the time of the earthquake to safe areas. These areas can be designated open spaces such as parking lots, public gardens and stadiums, as well as spontaneous shelter areas, such as areas between the buildings that are rubble free. In the present model, only designated open spaces are considered as safe areas. This choice was made to evaluate whether the city's land use planning sufficiently takes into account the space needed for emergency shelters.

However, several factors might affect the mobility of pedestrians and their arrival to open spaces. The first factor included in this study is the topography of the city: steep uphill slopes significantly reduce the speed of the individual. The second constraint considered is the presence of debris around the buildings, which slows pedestrians' movement. An additional considered criterion is the accessibility of open spaces. Indeed, physical gates could be found around an open space, that when locked, make the open space inaccessible, and transform it into a barrier.

### 2.2 Agents description

The architecture of the model is shown in Fig. 1. The model consists of four classes: open spaces, buildings, debris zones and persons. A class is by definition a template for an agent. When the model is implemented and the simulator is run, many agents for each class can be created following this template. For instance, many person agents that all have the same attributes are created, but with possible different values for the attributes, thus creating a heterogeneous artificial society.

**Open space agents**

This entity represents the open spaces in the study area. Open spaces are the target destinations of the evacuees. However, when they are locked, they are inaccessible. Each open space is characterized by its area, from which a maximum capacity is inferred by assuming a maximal density of 2 persons per square meter [21]. Whenever a person reaches an open space, the occupancy of this open space is updated, and the person's arrival time is added to the list of arrival times. The values of minimum, maximum and average arrival times of pedestrians in the open space are also updated.

**Building agents**

This agent represents the buildings in the city. All buildings are assumed to be residential. Each building is characterized by a number of floors and a number of apartments per floor. The building has a door, a point on one of its edges, from which the inhabitants can exit. It also has attributes related to its damage level: a damage state and an indoor death rate, that depend on the building's vulnerability.

**Debris zone agents**

This agent represents the danger area created by the falling debris around each building. Its main attribute is the casualty rate for a person that is located in this zone, which is dependent on the building's vulnerability and damage severity.

**Person agents**

This agent represents the people affected by the earthquake. In the current model, the individuals only react to their environment; they perceive their surrounding and react to it. Moreover, no social vulnerability characteristics are included at this stage: all individuals are considered to be healthy young adults with no disabilities, with more or less the same natural speed (i.e. speed on flat ground). An assumption that all individuals know where the open spaces are located and that they adopt the same behaviour and immediately head to the nearest open space, was also made. Nevertheless, if a person is located in a building on a higher floor, their evacuation will be delayed by the time it takes them to reach the building's exit.

The state of the person can be one of the following: in danger if they are in a debris zone, vulnerable if they are not in a debris zone but also not in an open space, and finally safe when they are in an open space. During the earthquake shaking, people might die inside the building, or if they are in a debris zone, where they might be hit by falling debris.

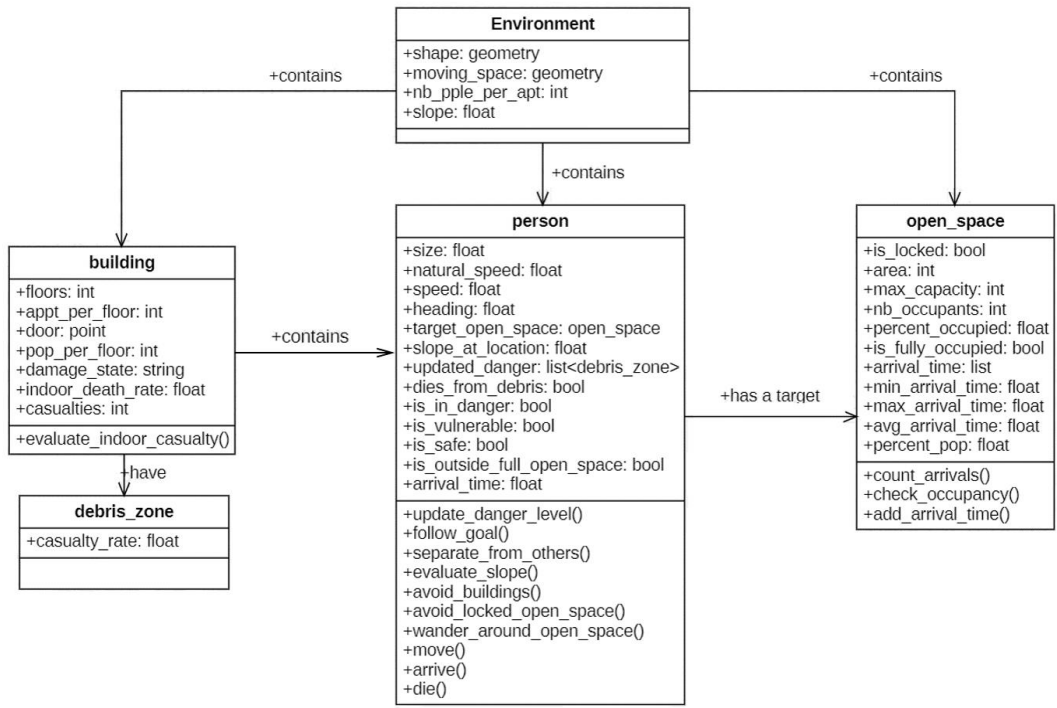

Fig. 1 – Simplified UML class diagram of the model

### Activity diagram of the person agent

The activity diagram in Fig. 2 depicts the behaviour of the person agents with the different situations and actions that can be encountered.

The agent chooses the open space closest to its location as target open space. At each simulation step, the person keeps moving towards their target location until they reach it. While moving, collision with buildings and other persons is avoided. The displacement of the agent is not constrained to the road network: it can move in the continuous space between buildings. The speed of a person agent can be modified by two constraints: (1) If it is in a debris zone, its movement is constrained by the presence of rubble. It was assumed that the agent's speed is reduced to half of its natural speed in this case. (2) The terrain slope affects the person's speed. Greater energy would be required to travel uphill. Similarly, a person would move slower downhill to ensure their safety against gravity. The relationships between the speed and the slope used in this model are extracted from [22].

Once the agent moves, if it is not in an open space, nor in a debris zone, its state is set to vulnerable. If it is in a debris zone, its state is changed to in danger. Additionally, if the earthquake has not yet ended, rubble might still be falling into the debris zone so the agent evaluates the probability of dying in this zone and might as well die at this given probability.

When an agent arrives to its target open space three situations might occur: (1) If the open space is locked, the agent changes target open space. (2) If the open space is accessible but has reached its maximum capacity, the agent keeps moving around this open space even if it is not safe. This is based on the assumption that a person would prefer to be close to the safe space and the other evacuees that have previously arrived, instead of travelling alone to another open space. (3) If the open space is accessible and has not reached its maximum capacity, the agent stays at the open space, and its state is set to safe.

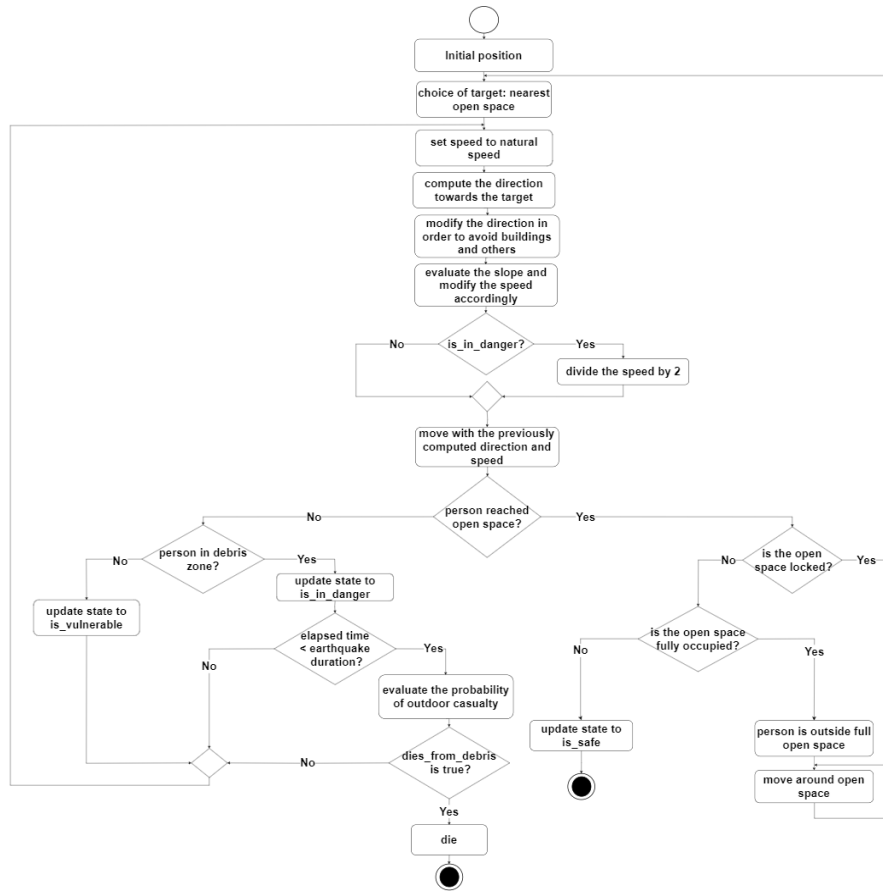

Fig. 2– Activity diagram of the person agent

## 3. Case study: Beirut, Lebanon

The model is applied to the city of Beirut, Lebanon. Lebanon has been historically subject to several strong earthquakes. The most remarkable are: the 551 AD (Mw 7.5) earthquake that was followed by a tsunami, the 1202 earthquake (Mw 7.5) that caused heavy destruction between the Lebanese coast and western Syria. The seismic event of 1759 (Mw 7.4) was the latest Mw>7 earthquake in Lebanon [23, 24]. Beirut is the capital of Lebanon and its political and economic centre. It is located on a rocky cape at the midpoint of Lebanon's Mediterranean coast. The city's relief is characterized by two hills: Moussaytbeh to the south and Achrafiyeh to the east. Beirut has witnessed in the last couple of decades a rapid urban growth in response to an increase of its population. This has led to a decline in open and green spaces [25]. This urbanization comes along with a lack of adequate urban planning and seismic provisions enforcement: the seismic regulation on buildings was introduced in 2005, but only made mandatory in 2012. The combination of all these factors make Lebanon, and Beirut especially, particularly vulnerable to earthquakes.

### 3.1 Data processing

This section describes the inputs required for the model. It also highlights the different data collection and preparation methods that preceded the implementation of the model.

**Earthquake scenario**

The decree 7946 introduced in 2012, considers Lebanon as one seismic zone. It fixes the minimal Peak Ground Acceleration (PGA) for which structures should be designed to 0.25g. However, an increase in the design PGA to 0.3g on Lebanon's coastal cities was proposed in [26], to comply with the recently discovered

Mount Lebanon Thrust fault. In this study, the latter proposition was retained to investigate the outcomes of this case. The scenario is fixed to an earthquake of PGA 0.3g. However, the earthquake scenario being only defined by the PGA, with no precision on the magnitude or the distance from the fault, a realistic strong motion duration could not be computed [27]. Thus, the duration of the shaking is hypothetically set to 30 seconds. The earthquake is assumed to be occurring at night where only the residents of Beirut are in the city. This was done in order to limit the uncertainties on the estimation of the population, since Beirut holds numerous institutional and business activities and is subject to a heavy migration flow during the day.

**Building data set**

Geospatial information on buildings in Beirut were collected from a previous project: the LIBRIS ANR (2010-2014) [3]. This project carried out field surveys to collect data on the buildings in Beirut. The construction year, the number of floors and the number of apartments per building are a few examples of the information collected. The buildings and their characteristics were digitized into GIS vector files, which were made available for the current work.

**Damage estimation**

To estimate the damages to the buildings, the vulnerability class of the buildings first had to be defined. The adopted classification relies on the construction year of the building and its number of floors, along with hypotheses on the evolution of construction practices in Lebanon [5]. The buildings were classified according to RISK-UE building typologies [4]. The classification was as follows: (i) For buildings built before 1950 with less than four floors the typology was assigned to be masonry. (ii) For buildings built before 1950 with four floors or more, and for buildings built between 1950 and 2005, their typology was assigned to be non-designed reinforced concrete. (iii) For buildings built after 2005, their typology was considered to be reinforced concrete with low ductility class.

The mean damage ($\mu_{DS}$) [4] of the buildings were computed using artificial neural networks that were previously trained [28]. These neural networks estimate the mean damage from simple indicators describing the level of seismic shaking and the vibrational properties of the soil and the building. The best neural networks performances are achieved with the following indicators: the PGA, the Peak Ground Velocity (PGV), the amplitude of the soil's H/V peaks ($A_{0H/V}$) and the ratio between the building's and site's resonance frequencies ($f_{bat}/f_0$). The obtained mean damages are on a scale from 0 to 4, which were then transformed to damage states ranging between: None, Slight, Moderate, Extensive and Complete.

**Indoor and outdoor casualty rates**

Casualty rates caused by buildings damages for indoor and outdoor situations are provided by the Hazus Manual [29]. Four casualty severity levels are defined, with severity 4 referring to a person being killed. In the present work, only the latter was considered. Therefore, for each building, the corresponding indoor and outdoor casualty rates were retrieved, depending on its typology and damage state.

**Debris estimation**

The buildings in Beirut being mostly constructed in masonry or in reinforced concrete frames with masonry in-fills, debris are expected during an earthquake. Therefore, the footprint of the debris should be taken into account in the model. From observations and analysis of the usual patterns in buildings damage, it was assumed that a rectangular building collapses in a truncated pyramid shape [30]. The truncated pyramid is made up of debris, which is formed by the building's construction material. The calculation of the debris area around a building is therefore summarized to finding the values of $x_p$ and $y_p$, the dimensions of truncated pyramid's base (Fig. 3). As a first step, an equivalent rectangular shape for each building was calculated. The rectangle has an area and a perimeter equal to the original building's shape. Before collapse, the height of each floor is assumed equal to three meters, thus the total height of the building is:

$$h = 3 \times n \tag{1}$$

$n$ being the number of floors.

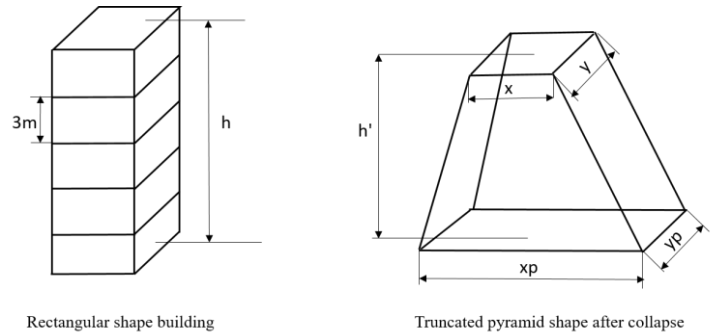

Fig. 3- Building's shape before and after collapse

The building is considered to lose one meter per floor during collapse, so the height of the truncated pyramid was calculated as:

$$h' = h - n \qquad (2)$$

We assumed that the construction material takes up one third of the original building's volume. In case of partial collapse, the volume of generated debris is proportional to the normalized mean damage:

$$V = \left(\frac{1}{3}\right) \times A \times h \times \mu_{NDS} \qquad (3)$$

Where A represents the area of the building and $\mu_{NDS}$ represents the normalized mean damage. *V* thus represents the volume of the truncated pyramid, and may be expressed as the difference in volume between the total pyramid and the top part of the pyramid (the removed part):

$$V = \frac{x_p \times y_p \times h_t}{3} - \frac{x \times y \times (h_t - h')}{3} \qquad (4)$$

With $h_t$ being the total height of the pyramid. By using similar triangles and Pythagoras' theorems, the following equations were obtained:

$$\left(\frac{x}{x_p}\right)^2 = \frac{(h_t - h')^2 + \frac{y^2}{4}}{h_t^2 + \frac{y_p^4}{4}} \qquad (5)$$

$$\frac{x}{x_p} = \frac{y}{y_p} \qquad (6)$$

Finally, $x_p$, $y_p$ and $h_t$ were found by the solving equations (4), (5) and (6). The debris zones were then created by applying a buffer distance around each building equal to the maximum between ($x_p$ - x) /2 and ($y_p$ - y) /2.

**Digital elevation model generation**

To derive the elevation and the slope of the ground in the city of Beirut, three Pleiades satellite images with a resolution of 0.7 meters acquired over Beirut in July 2016 were used. These tri-stereo images were correlated using Ames Stereo Pipeline [31] which generated a digital surface model (DSM) of 2 meters resolution. A DSM represents the elevation of the earth surface along with the elevation of the other features above the ground, such as trees and buildings. However, for this study the topography of the natural terrain without the elevated features was needed. So additional steps had to be achieved to find the bare terrain's elevation. The DSM was introduced in Quantum GIS (QGIS 3.4.4-Madeira). The SAGA-GIS module DTM Filter (slope-based) was used to classify the DSM's cells into bare earth and object cells (ground and non-ground cells). The bare earth cells were interpolated using the Multilevel b-spline method which produced an elevation model of the natural terrain. The slope of the natural terrain was then calculated in QGIS and saved into a separate vector layer.

**Population estimation**

An accurate estimation of the population's size and distribution is an important factor to recreate a realistic spatial representation of Beirut's' residents. However, in developing countries demographic data is frequently unavailable due to the absence of the mechanization of population system [32]. This is the case for Lebanon, since the last national population census dates back to 1932, and no official population estimations have since been undertaken. To counter the lack of data, a result from a survey carried by the Central Administration of Statistics (CAS) was considered as reference [33] and applied to the building dataset. This study investigated the average household size in Beirut, which was found to be equal to four. The residents in each building were estimated by multiplying the number of apartments in the building- provided by the buildings dataset- by the average household size. This gives an estimation of the number of people in Beirut for a night-time scenario.

## 4. Simulation and results

The GAMA platform was chosen for the simulation of the model. GAMA is an open-source modelling and simulation platform that allows the development of multi-agent simulations in a spatially explicit environment. It offers an intuitive modelling language, GAML, and a large library of toy models that can serve as a starting base for the implementation. One of the main advantages of GAMA is its ability to handle realistic spatial dataset and large-scale simulations with thousands of agents, which is a crucial factor for city-scale simulations.

### 4.1 Simulation

The simulator is still in the testing phase. The simple assumptions taken in the model are tested, and once validated, the complexity of the model will be progressively increased. A zone in the district of Hamra in Beirut was used to run the first simulations (Fig. 4). 745 buildings are located in this area of 0.885 km$^2$. Most of these buildings are slightly damaged (81%) for the considered earthquake scenario, while 18% are moderately damaged (Table 1). 37 750 people are estimated to live in this zone. We lack the information on designated open spaces; therefore, 8 open spaces were digitized at construction-free locations in the study area.

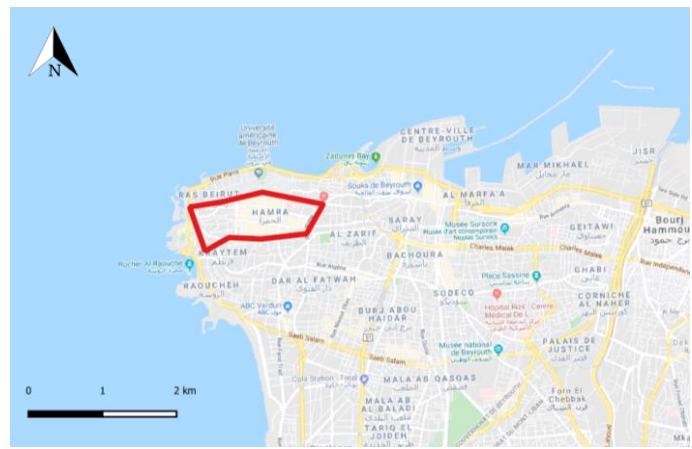

Fig. 4 – Snapshot of Beirut's map. The simulation zone is displayed in red. Source: Google Maps.

Table 1 – Distribution of the damage states corresponding to the buildings in the test zone

| **Damage state** | **None** | **Slight** | **Moderate** | **Extensive** | **Complete** |
|---|---|---|---|---|---|
| **% of buildings** | 0.8 | 81 | 18 | 0.20 | - |

The GIS data related to buildings, debris zones, open spaces and topography were introduced into GAMA. The corresponding agents were directly created from these files. On the other hand, the person agents were gradually created throughout the simulation at the door of each building. Fig. 5 shows a 3D view of the simulation environment in GAMA at cycle 45 (time=45 s). Buildings are coloured according to their damage state: buildings with Slight damages are coloured in Green, and those with Moderate damages in yellow. The debris zone around buildings is coloured in salmon. Open spaces are numbered from 1 to 8 and they are coloured in light blue if they are open, and in dark blue if they are locked. People are displayed in red if in danger, in orange if vulnerable and in green if safe. Two city configurations were simulated. Configuration 1 is a configuration where all the open spaces are accessible. Configuration 2 is where one open space is locked while all of the other open spaces are accessible. Open space 1 was set as locked in this configuration, because of its position in the centre of the study area. The duration of the simulations is 5 minutes, with each simulation step representing one second.

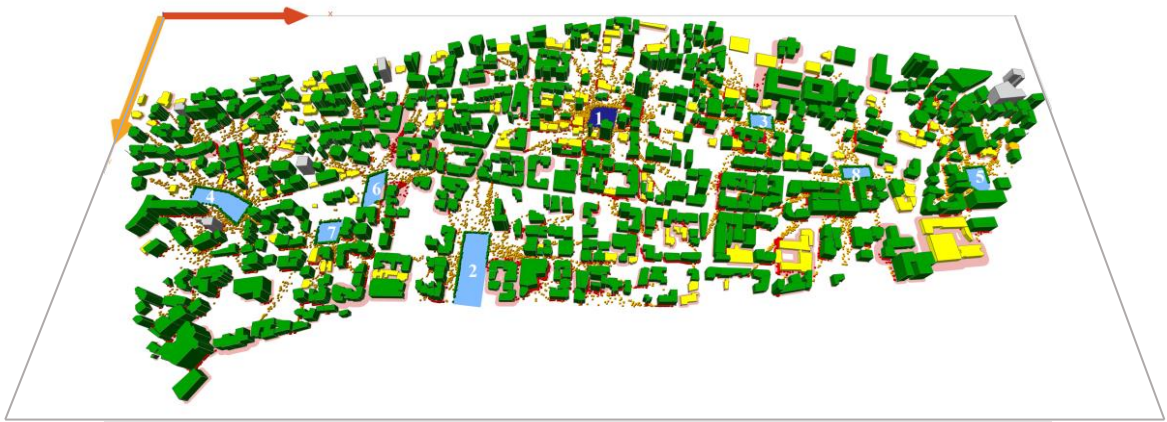

Fig. 5– 3D snapshot of the simulation environment in GAMA at time=45s for city configuration 2.

### 4.2 Results

Fig. 6 shows a graph of the number of safe people at each simulation step, for the two city configurations. The number of people arriving in open spaces increases with time, and reaches 19 469 (52% of the population) within 5 minutes in the case where all open spaces are accessible. However, when only one open space is locked, open space 1 in configuration 2, only 39% of the initial population is able to reach an open space within 5 minutes. This observation could be explained by the fact that people would have to head towards a further open space when the open space near their residence is inaccessible, which delays their safety. These results show that the presence of accessible open spaces in a city and their proximity to the residential buildings is a crucial factor for ensuring people's safety when an earthquake occurs.

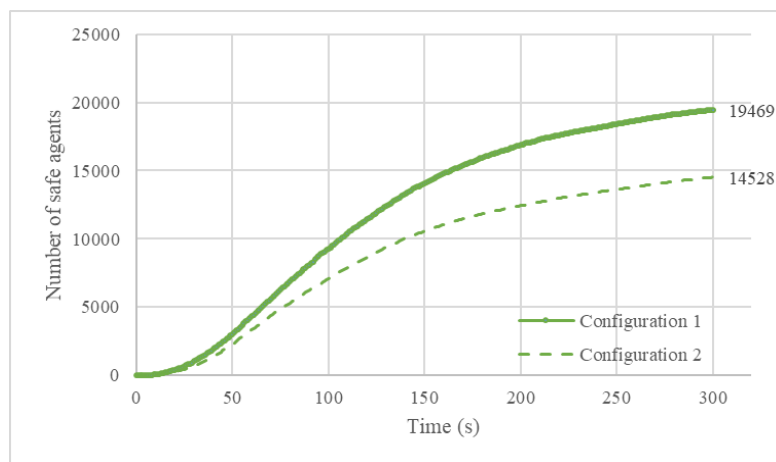

Fig. 6– Number of safe people agents at each simulation time step for city configurations 1 and 2.

Table 2– State of open spaces at the end of the simulations for city configuration 1 and 2 [between brackets].

| Open space | #1 | #2 | #3 | #4 | #5 | #6 | #7 | #8 |
|---|---|---|---|---|---|---|---|---|
| % population | 13 [0] | 7 [6] | 7 [6] | 8 [9] | 7 [7] | 4 [5] | 3 [3] | 3 [3] |
| Minimum arrival time (s) | 4 [0] | 6 [6] | 5 [5] | 3 [3] | 4 [4] | 8 [7] | 13 [14] | 7 [6] |
| Maximum arrival time (s) | 300 [0] | 300 [300] | 300 [300] | 300 [300] | 300 [300] | 300 [300] | 299 [300] | 297 [300] |
| Average arrival time (s) | 122 [0] | 117 [119] | 113 [113] | 114 [113] | 130 [129] | 115 [115] | 122 [121] | 79 [78] |

The population percentage in each open space at the end of the simulations, along with the minimum, maximum and average arrival times for each open space, in the two city configurations, are shown in Table 2. In terms of arrival times, similar trends are observed for configurations 1 and 2. The maximum arrival times in all open spaces are around 300s, which means that no open space was full in the 5 minutes that followed the earthquake. Additionally, the average arrival times are around 120 s for all open spaces, except for open space 8, where the average is around 80s. This could be explained by an important number of arrivals within the first minute, and a decrease in the flow of arrivals at later stages. As for the population percentage, the distribution of the population in open spaces in configuration 2 is almost similar to that in configuration 1, except for open space 1 that was not accessible. Moreover, the difference in the number of safe people in configurations 1 and 2 (13%), is equal to the percentage of the population in open space 1 in configuration 1. This means that, in configuration 2, the people who headed towards open space 1 and found out it was locked upon their arrival, could not arrive to any other open space within 5 minutes. This observation shows that longer simulations should be run in order to observe the emergence of a different population distribution in open spaces.

## 5. Conclusion

In this paper, an interdisciplinary seismic risk model integrating the human behaviour is presented. Pedestrians' evacuation in a city, during and a few minutes after an earthquake, is modelled using an agent-based model. The model simulates the individuals' mobility in an environment that has been modified by an earthquake, with the constraints that affect it. These constraints are assumed to be: the topography, the debris generated around the buildings, and the access to open spaces. Therefore, the model integrates geospatial information on the city's buildings and topography, as well as realistic estimates of damages and debris footprints.

This model was applied to the city of Beirut, Lebanon that displays a high vulnerability to earthquakes due to its historic seismicity and buildings vulnerability. The considered seismic scenario is an earthquake with a PGA of 0.3g and a duration of 30 seconds occurring at night. Therefore, only the residents of the city are included in the simulations. Simulations of 5 minutes from the beginning of the earthquake are performed for two city configurations: (1) with all open spaces accessible and (2) with one of the open spaces locked. The results show that 52% of the population is able to arrive to open spaces within 5 minutes when all open spaces are accessible. This rate is reduced to 39% when only one open space is considered to be locked.

Future work will focus on revising the simplification assumptions taken in the model, to render it closer to reality. This could include: (1) including social vulnerability attributes to the pedestrians to represent different population characteristics, (2) adding cognition and decision-making abilities to the pedestrians, (3) including spatial information on buildings and open spaces extracted from satellite images, (4) taking into account different buildings uses and the distribution of the population in the city at different times of the day. As for the simulations, future works will focus on: (5) simulating up to 15 minutes after an earthquake, (6) running simulations on an increased geographic extent to cover the whole city of Beirut.

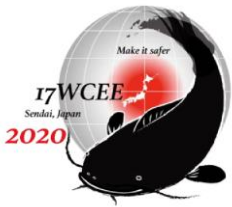

## 6. Acknowledgements

This work is supported by the French National Research Agency in the framework of the Investissements d'Avenir program (ANR-15-IDEX-02). It is also partially supported by the University Agency of the Francophonie in the framework of the Scientific Inter-university Collaboration Project program. The authors acknowledge the contributions of the LIBRIS ANR project (2010–2014) funded by the French national Research Agency, and the Lebquakechat project (2019) funded by French National Center for Scientific Research.